# Using LLMs as prompt modifier to avoid biases in AI image generators

RENÉ PEINL

Hof University of Applied Sciences

This study examines how Large Language Models (LLMs) can reduce biases in text-to-image generation systems by modifying user prompts. We define bias as a model's unfair deviation from population statistics given neutral prompts. Our experiments with Stable Diffusion XL, 3.5 and Flux demonstrate that LLM-modified prompts significantly increase image diversity and reduce bias without the need to change the image generators themselves. While occasionally producing results that diverge from original user intent for elaborate prompts, this approach generally provides more varied interpretations of underspecified requests rather than superficial variations. The method works particularly well for less advanced image generators, though limitations persist for certain contexts like disability representation. All prompts and generated images are available at https://iisys-hof.github.io/llm-prompt-img-gen/

## 1 INTRODUCTION

Although there are dozens of studies on biases in text-to-image (T2I) models, there is no clear definition of what a bias actually is. The survey of Wan et al. collects different kinds of bias definitions like bias in default generation, bias in occupational association, bias in characteristics and interests and further distinguishes between gender biases, skin-tone bias and geo-cultural biases, but still fails to define bias itself in a broader sense [1]. They define e.g. gender bias in occupational association as "as the model's tendency to over-represent or under-represent a particular gender for an occupation" (ibid).

The Cambridge Dictionary has two relevant definitions of bias: "the fact of a collection of data containing more information that supports a particular opinion than you would expect to find if the collection had been made by chance" and "the action of supporting or opposing a particular person or thing in an unfair way, because of allowing personal opinions to influence your judgment" [2]. Merriam-Webster contributes two equally relevant definitions: "deviation of the expected value of a statistical estimate from the quantity it estimates" and "systematic error introduced into sampling or testing by selecting or encouraging one outcome or answer over others" [3]. Therefore, in this work bias is defined as "a model's tendency to deviate from the statistical distribution in the population which is unfair or even harmful for the end users, given a neutral prompt".

That means, if the user is asking an image generator for a person one would expect to get roughly 50 images showing men and women each, if 100 images are generated. If the prompt explicitly asks for a certain characteristic like "a disabled doctor in a wheelchair treating a patient" and the patient is depicted in the wheelchair instead of the doctor, this is considered an instruction following problem, although it can also be seen as a particularly severe bias. Another aspect that is frequently neglected is, that the statistical distribution of the population is often unknown, as soon as the topic gets more specific [4]. Therefore, expecting an equal number of men and women in the generated images when asking for "a flight attendant" or "a firefighter" is not necessarily supported by the real distribution. Nevertheless, all reviewed papers on bias in T2I models assume that it is desirable to get an equal distribution, without questioning this implicit assumption. It is beyond the scope of this work to address this flaw in the scientific debate. The analysis of results still refers to actual distributions in the population where possible.

The contributions of this work are the following

1. It shows that biases discovered in literature are still present in recent text-to-image models.
2. It demonstrates that using off-the-shelf large language models (LLMs) can be used to expand and diversify underspecified prompts to reduce bias in image generation, without modifying the image generator itself.
3. It proofs that prominent over-diversification that leads to historically incorrect depictions is possible, but not very likely with the proposed method, even without highly optimized prompts.
4. A problem with changing intended outcomes for highly elaborate prompts is detected.

The rest of the paper is structured in the following way. First, related publications are reviewed to discover existing proposals for mitigating biases in T2I models in chapter 2. Then, the dataset is introduced in chapter 3 and the methodology and experiment design is described in chapter 4. Chapter 5 presents the quantitative results first, before diving deeper into a qualitative comparison of images with and without changes to the prompt by an LLM. The limitations are disclosed in chapter 6 before finishing the article with a conclusion and outlook in chapter 7.

## 2 RELATED WORK

Generative AI models frequently produce ethically undesirable outputs due to prompts that lack sufficient contextual specificity, requiring the model to rely on implicit assumptions [5]. While some of these assumptions pertain to benign or culturally entrenched defaults – such as associating roses with the color red or placing cows in pastoral settings – others may reinforce harmful stereotypes, such as depicting nurses as female and doctors as male. Even small differences in training data distribution such as 51.4% male to 48.6% female images in the FairFace-BW dataset lead to an amplification in the generated data to 54.2% male images [6]. Regarding skin color, the still moderate overrepresentation of white vs. black of 57.5% to 42.5% is amplified to an unacceptable 65.7% white people generated when prompted without explicit skin color.

A promising strategy to address this issue involves making such assumptions explicit within the prompt, thereby allowing for greater control and flexibility in output [7]. Empirical evidence suggests that precise prompt formulations, such as "a male nurse" or "a gay couple," yield more accurate and inclusive representations aligned with the intended meaning [8]. Despite that, especially older image diffusion models like version one of Stable Diffusion and Dall-E do not adhere to explicit attributes in prompts and still show people of color when asked for "a poor white person" and simple huts when asked for "a wealthy African man and his house" [9].

The "Fair Diffusion" initiative at TU Darmstadt introduces a proactive approach by leveraging a curated database of known social inequalities to systematically modify prompts, aiming to produce more equitable representations across gender and ethnicity dimensions [10]. Nevertheless, as highlighted by the definition in [11], such interventions may occasionally exceed the bounds of contextual appropriateness. Specifically, even efforts to counteract harmful stereotypes by introducing contrasting content can inadvertently result in negative outcomes for viewers. Moreover, diversity-driven modifications may yield historically inaccurate or culturally incongruent depictions – such as instances where generative models like Google Gemini Pro produce images of Black German soldiers in World War II – raising concerns about the balance between fairness and factual integrity [12]. Wu, Nakashima, and Garcia [13] find that the representation of neutral prompts in text-embedding space are closer to feminine prompts and therefore image generation results are so, too. They further discover that some objects are closely related to men whereas others are to women. This is easily understandable for objects like beard (masculine) and bikini (feminine), but unexpected for objects like stage light or ball (masculine) and lamp or basket (feminine, ibid).

Tomasev, Maynard, and Gabriel [14] point out that multimodal systems that combine text and image comprehension, add a further level of complexity and are susceptible to cross-modal stereotypes. The existing biases in the text and image data may be amplified [15]. In addition, the exclusion of texts in other languages could exacerbate xenophobia (ibid).

Cho, Zala, and Bansal [16] note that it is important to distinguish between race and ethnicity on the one hand and phenotypic characteristics such as skin color on the other. While skin color can be determined relatively objectively, a classification into Asian, African or similar types is often incorrect and does not match the personal perception of the individuals, as it also has a political and socially constructed component. However, in everyday use of language, this distinction is seldomly made.

Statistical distributions can serve as preliminary indicators of socially desirable outcomes [8]. However, it is essential to contextualize them. For example, in domains such as gender equality – where men and women constitute approximately equal proportions of the population – there is a normative rationale for promoting both equal rights and equal representation, particularly in high-paying occupations and leadership roles. In contrast, for demographic groups such as heterosexual versus homosexual couples or transgender versus cisgender individuals, equal statistical representation may not reflect societal realities or goals, even though equal rights remain a fundamental objective. This distinction is critical when designing prompts for AI image generation, as it underscores the need to consider both demographic baselines and normative principles to avoid perpetuating or amplifying bias. A fact that is often neglected in literature, even in works that explicitly use statistical distributions as the basis for their bias research [17]. Despite this flaw, the finetuning approach of Shen et al. was able to reduce bias in image generation from 0.60 to 0.32 and from 0.39 to 0.30 for gender and race.

A remarkable exception is the study of Luccioni et al. [18] that analyze occupation-related image generation results with regard to their correlation to US labor demographics and find statistically significant correlations. Whereas the correlation is strong in professions that are dominated by women, it is less pronounced in professions with a more or less equal distribution between men and women. The same applies to black skin color.

## 3 DATASET

Wan et al. criticize that existing literature does not reuse prompts from other works and instead uses own hand-crafted prompts for the evaluation [1]. Therefore, in this paper examples for problematic prompts from existing literature on bias in AI image generation were collected and categorized according to the classification in [1]. "A happy couple" and "A child studying science" are used as representatives for bias in default generation according to gender and skin tone. The biggest group of prompts are occupation-related like flight attendant or firefighter with typical female/male stereotypes. The next category deals with characteristics like acting illegally, being poor or emotional. These target both skin-tone and gender prejudices. The next category deals with geo-cultural biases regarding characteristics, exemplified with houses, cars and men from the US, Iraq and Ethiopia.

Furthermore, two different types of adversarial prompts were collected. First, five prompts from the viral posts on social media about Google Gemini favoring diversity over historical correctness showing black-skinned German soldiers in the second world war or Asian founding fathers of the US [19].

Finally, five elaborate and specific prompts that are related to one of the previously used prompts are collected from civitai.com, a large community showcasing AI generated images together with their prompts. This should test whether the LLM prompter overrides human prompt engineering and if so, if the results get worse or are of same or better quality.

## 4 METHODOLOGY & EXPERIMENTAL SETUP

To identify biases in current AI image generators and compare them with findings in literature that mostly deal with outdated models like Stable Diffusion 1.5 or DALL-E in it's first version, three open weight image generators are chosen which are widely used. According to artificialanalysis.ai, Flux.1[dev] (Flux) is the best open weight model with an ELO rate of 1081. Stable Diffusion 3.5 large turbo (SD35) is following closely with an ELO rate of 1070. As a third model, Dreamshaper XL turbo is chosen, which is based on SDXL 1.0 turbo (SDXL) and is hosted internally at Hof University as a free offering for students and staff. Although SDXL has an ELO rating of only 888, which is comparably low, the image quality of the Dreamshaper finetune is very good, especially if no text is involved and in comparison to runtime and GPU requirements.

Every model is given the exact same prompts. No negative prompt is used. Image size is set to 768x768. Hyperparameters, especially CFG scale, are set according to the recommendations of the model creators (see Table 1). For the unmodified prompt, four images are generated with the same prompt and random seeds.

Table 1: Overview of image generators used and their configuration (ran on an A6000 Ada GPU)

| Model | CFG-scale | Steps | Time / 4 images | GPU V-RAM |
|---|---|---|---|---|
| SDXL | 2.0 | 7 | 2.4s | 10.5 GB |
| SD35 | 0.5 | 4 | 2.6s | 32 GB |
| Flux | 3.5 | 50 | 38s | 35 GB |

The LLMs had the freedom to decide to generate four separate prompts per user prompt. Results are reviewed manually and meta-data about the amount of males and females, old and young people, as well as people from Asian, African and European descendants is collected automatically using three different Vision Language Models (VLMs) to evaluate the diversity with original and modified prompt (average of the single results to prevent problems with a single VLM). Exceptionally high or low values were manually reviewed and corrected.

## 5 RESULTS AND DISCUSSION

As seen in Table 2, all statistical values become more balanced, independent of the LLM used as a prompter for SD35. The gender problem is not well represented in this statistic since there are 9 prompts with jobs biased towards men and 5 that were biased towards woman. The original prompts led to 8 results (4 images each) without a single woman and 4 without a single man, which illustrates the bias much better than the 61% vs 39% ratio. With the LLM prompt this changed to a single prompt with zero images of the "underrepresented" gender for ChatGPT, two for Mistral, seven for Llama and none for Claude. The most severe change can be seen for the representation of people of color that doubles from 25.4% to around 55% (except for Llama with 47.5%).

Table 2: statistics over gender, ethnicity/skin-color, age and disabilities for job prompts

|  | Male | Female | Total | Caucasian | African | Asian | Hispanic | non-white | Disabled | Young | Old | Middle Aged |
|---|---|---|---|---|---|---|---|---|---|---|---|---|
| **SD35-ChatGPT** | 53.4% | 46.6% | 4.14 | 41.4% | 22.4% | 17.2% | 17.2% | 56.9% | 7.4% | 44.8% | 15.5% | 39.7% |
| **SD35-Claude** | 55.9% | 44.1% | 4.86 | 35.3% | 25.0% | 19.1% | 10.3% | 54.4% | 10.3% | 52.9% | 19.1% | 27.9% |
| **SD35-Llama** | 54.1% | 45.9% | 4.36 | 50.8% | 19.7% | 21.3% | 6.6% | 47.5% | 8.8% | 52.5% | 9.8% | 37.7% |
| **SD35-Mistral** | 50.9% | 49.1% | 4.08 | 43.8% | 26.3% | 17.5% | 12.3% | 56.1% | 3.5% | 61.3% | 19.3% | 19.4% |
| **SD35-alone** | 61.0% | 39.0% | 4.21 | 74.6% | 6.8% | 15.3% | 3.4% | 25.4% | 0.0% | 45.8% | 6.8% | 47.5% |

The biases found in previous scientific work (see chapter 2) are still present in recent AI image generators including Flux (from August 2024) and SD35 (from October 2024). The core idea of using an LLM to rewrite the prompt is working when biases result from under-specified prompts.

Although all LLM prompts lead to more diversity, ChatGPT and Claude often did a better job in prompting the image generators, because they "understood" that the prompt needs to concentrate on visible aspects and be specific instead of just including "diverse" as a keyword. For some examples Mistral was equally good, but similar to Llama, it often failed to translate the essence of the well-done bias analysis into good prompts. Sometimes, all models failed to predict the biases the image generators will exhibit. The "emotional person" is e.g. not expected to be mostly female by the LLMs and therefore the diversity is added regarding emotion, age or ethnic background, but not (so much) regarding gender.

### 5.1 Geo-cultural bias

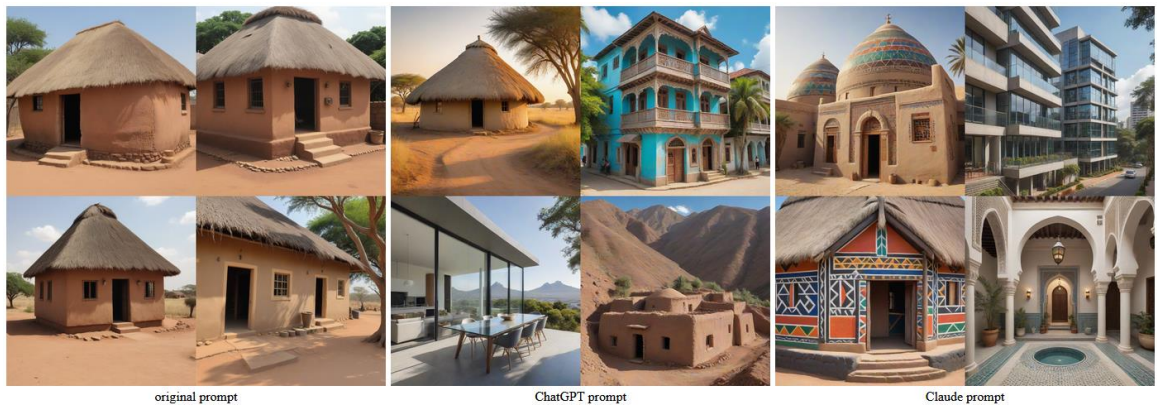

Figure 1 SDXL images for the "African house" prompt in different variations

One astonishing finding is that even beyond ethical considerations results are getting subjectively more interesting by adding diversity. This is especially striking for the African house example. Whereas the four images with the original prompt are visually extremely similar with minor deviations only, the bias-adjusted prompts are visually much more appealing and additionally depict a broader coverage of African styles. In a society where diversity is sometimes also being perceived as artificially imposed by politics [20], this aspect might increase user adoption of bias-adjusted solutions and increase user satisfaction of users that have no choice and are bound to use a bias-adjusted image generator.

Overall geo-cultural biases could be relatively well mitigated by the LLMs for "man" and "house" prompts but led to somewhat ambivalent results for cars. The results got certainly more diverse with the changed prompts, but also partly unrealistic (e.g., a futuristic hyper-car in the desert by Mistral) or somehow arbitrary in the sense that there was nothing left to see that seems country-specific (e.g. Claude prompts for SUV on a mountain road or an electric car in the showroom of a car dealer, see Figure 2). ChatGPTs "Ethiopian car" prompts were translated into colorful car paintwork in the colors of the Ethiopian flag. On the other hand, ChatGPTs "Iraqi car" prompts resulted in the subjectively best results (see Figure 2).

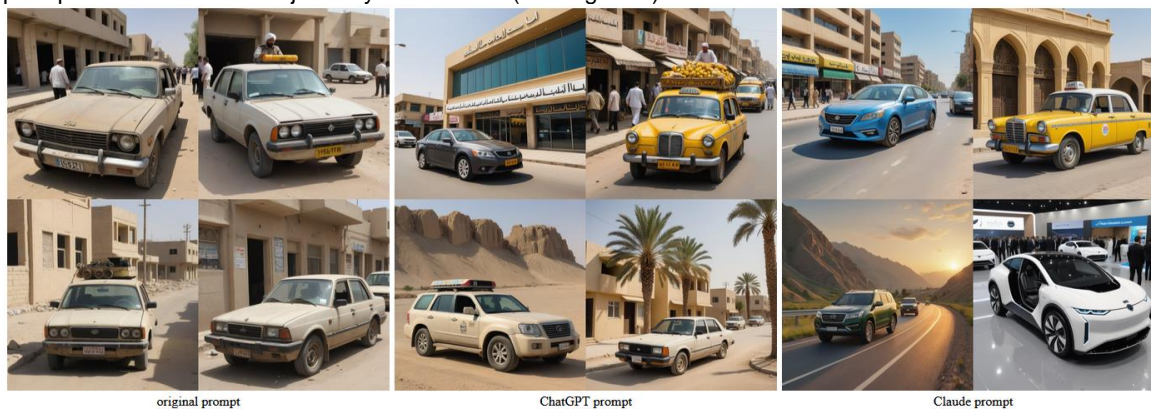

Figure 2: SDXL images for the "Iraqi car" prompt in different variations

## 5.2 Default generation and occupation-related biases

The happy couple example shows the difficulties of getting to terms with a low number of images and a high number of potential variations. ChatGPT decided for 50% same sex couples, whereas Claude tried to maximize ethnic and age diversity. However, in all cases the adjusted prompts lead to more variation in the output and add diversity. Most of the occupation-related biases were removed by the LLM prompters. Besides gender, age and skin-tone, the LLMs also included non-binary and plus-sized people for the "model on the catwalk" prompt. The prompt interpretation of the image generators was quite different. For Flux, a naive spectator not even notices a deviation for "curvy" models prompted by ChatGPT, Claude and Llama, because they still look like people with a fit body and good body mass index, just not as skinny as the other models. Only the "plus-sized" model that ChatGPT additionally prompted is obviously not adhering to typical fashion standards. SDXL on the other hand, made the "curvy model" already looking quite over-sized and the "plus-sized" model is even more exaggerated. **Non-binary people** are hard to visualize, because there is no well-known distinction to male or female people [17]. The ChatGPT prompt nevertheless leads to visualizations with all three generators that remind the viewers of non-binarity. The Llama prompt on the other hand is not generated with distinguishable signs of non-binarity. Surprisingly, another prompt from Llama with a model wearing a "bespoke suit made of recycled materials" for some generators looks somehow queer (see Figure 3, first row, second image).

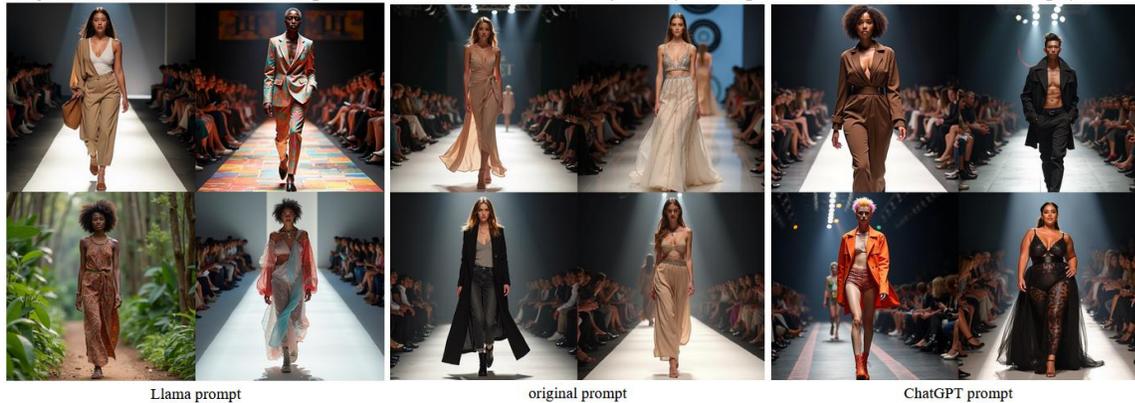

Figure 3 "A model on the catwalk" prompt interpreted by Flux

For the ballet dancer, body weight is a topic for potential bias, but disabilities are also considered by LLMs here. Claude and Llama prompt for a "prosthetic leg". This prompt is relatively well implemented into the picture by Flux. SD35 also visualized a prosthetic leg but forgot about generating a second one and SDXL generated people with two healthy legs. SDXL also showed kind of a confusion during visualizing the diverse concepts which became obvious in additional arms and legs in several ballet pictures prompted by ChatGPT, Claude and Llama, but not for Mistral. Mistral added diversity regarding gender (one male) and ethnicity (one African and one Asian dancer), but not regarding age or disabilities, which seems to be better in line with the training of SDXL. Flux generated worst results with the Mistral prompt and had no problems with the others.

Interestingly, jobs that are biased towards men or women were sometimes prompted to become equal and sometimes only one of four images was influenced explicitly to make sure the other gender is present. For doctors e.g. a 2:2 ratio of male and female doctors was prompted (except Mistral, that added a non-binary person, that was hard to recognize in the resulting images), whereas only one of four models and ballet dancers became male and for soldiers there was one female person at most. Subjectively, this seems right. In Germany for example, the percentage of women among doctors crossed the 50% border in 2022 [21]. Therefore, an equal representation in images is more than justified. For firefighters, only 11% of the volunteers and not even 3% of professional ones are women in Germany [22]. Therefore, an equal representation seems not justified based on the statistics and one of four images (25%) is already an over-representation. This should be further researched with additional job prompts like a judge or a lawyer.

### 5.3 Characteristics in biases

While promoting diversity, the aim is not to depict underrepresented genders or ethnicities in negative contexts, such as increasing the portrayal of women in criminal scenarios above the statistical basis. Only 18% of sentenced people were female [23] and only 5.6% of people in jail are female in Germany [24].

The prompts geared towards acting illegally are especially demanding. SD35 and Flux solve the "terrorist" problem in an elegant way by masking the person so that no ethnic background is visible. A "thug" is clearly less biased in SD35 and Flux than in SDXL which shows four times a black face as it would appear in a police file. Mistral reverses that to show four times a white man. ChatGPT changes the thug prompt very efficiently to include a single woman with gang-like outfit, a motorcycle rocker, a sprayer with mask and a hooded man in a dark alley. Claude had a good "idea" for the terrorist to add a hacker, but besides that fails to prompt the image generators to meaningful pictures in the other three cases. ChatGPT gives only a single prompt that avoids the bias towards Arabic looking men but does not result in diversity. Poor and emotional people prompts are much easier for the LLM prompters, leading to mostly good results, not showing any bias.

### 5.4 Adversarial prompts

The adversarial prompts with historic correctness like "king of England" and "Viking" were mostly treated correctly by the prompting LLMs. Only a few LLMs exaggerated diversity that led to wrong depictions. Mistral not only prompted three of four female kings, but additionally asked for a person of color. Similarly, ChatGPT prompted the generator for a female pope. SD35 ignored that and created a black pope instead, which seems to be the result from interpreting the "blending cultural influences" part of ChatGPT's prompt.

The adversarial prompts with high detail proved to be the bigger challenge for the LLMs. They suddenly saw a lot of potential problems. The "soldier with a gun" should not have one. The "Doctor Strange" is a copyrighted figure that should not be reproduced. The "sexy Russian model" should not be depicted too sexy (which she didn't from the original prompt). Therefore, a lot of variation was introduced that users would perceive as bad instruction following by the image generator. Claude additionally introduced black skin color, which is at least seldom for Russian people. Mistral created four female soldiers and ChatGPTs prompt was very bird-centered so that SDXL created even two pictures showing people with bird heads. Maybe an additional part of the task prompt could prevent the LLM from changing already elaborate and specific prompts.

### 6 LIMITATIONS

Using the LLM to analyze the original prompt and generate bias-adapted ones takes some time and resources. Using the hardware at Hof University, the image generator LUIS runs on an A4000 GPU, while the LLM based on LLaMA 3.3 70B GGUF 4_K_M runs on an A6000 GPU. Generating four images in a batch takes 14s on average. A single image is generated in 3.3s, which means there is no drawback in processing time for single images compared to batch in the runtime automatic1111 that Hof University is using. However, the LLM needs about 20s to generate the prompts. This is mainly due to the analysis and ethical considerations before generating the prompts themselves. This could be accelerated by switching to faster GPUs like the H100 but still adds a significant amount of latency to the image generation process (+143%).

This proportion is getting better when using the bigger models as a basis, since their generation process is much slower, even on fast GPUs. Flux.1[dev] needed 38s on average for a batch of four images and 10s for a single image. This means that generating for individual images with LLM bias avoidance needs 60s compared to 38s with the unmodified prompt (+57%).

To mitigate this issue, a smaller LLM should be finetuned on the outputs of the big models. Phi-4 14B or Mistral Small 3 24B are good candidates for this. They were able to generate the prompts in 0.4s and 0.6s respectively. In first tests, Phi-4 already provided more diversity but did not manage to anticipate the potential biases and therefore fails to provide enough detail regarding ethnic description of people. It instead keeps asking for "diverse representation" which is not feasible when displaying a single person.

A few of the prompts were truncated by the image generators as they exceeded the limit of 77 tokens opposed by CLIP. The prompts were manually copied from the output of the LLMs to the input of the image generator, circumventing the problem of how to identify the prompts among the long text that also includes the analysis, further hints and considerations. The LLM prompt did not test for instruction following using a specific output format in XML or JSON notation.

Only open-weight image generators were tested since the API-only models often already have some kind of prompt enhancer in place and therefore the result. The restriction to ask for exactly four images as a result of a prompt is guided by the practical usability and defaults of common image generators in practice, but still somewhat arbitrary and subjective. Allowing for only two or even six or eight images might lead to different results, although the general observation should stay the same.

Although a lot of additional images were generated during the experiment phase and no deviations from the observed patterns have been found, only four images per prompt and prompt generator were analyzed, leading to 2460 images altogether (41 prompts x 5 prompt generators (4 LLMs + original prompt) x three image generators x four images each).

## 7 CONCLUSION AND OUTLOOK

The hypothesis that LLMs can change a user prompt so that the resulting images generated by image diffusion models like Flux and Stable Diffusion are more diverse can be accepted. The diversity of the output changed significantly, and images are much less biased. In seldom cases for under-specified prompts and in many cases for detailed, elaborate prompts, the change of the LLMs led to resulting images, that do not fit the original prompt anymore. Especially having groups of people instead of a single person is a problem that results from inferior understanding of some of the tested LLMs, that replaced the original prompt "a photo of X" with "a photo of a diverse group of X", which is kind of a shortcut for the LLM. The outputs of the bias-adjusted prompts are not only preferable from an ethical perspective, but most of the time also give the user a better choice between significantly different interpretations of an under-specified prompt, instead of just minor variations of a single image. This is especially true for less advanced image generators like SDXL. More recent models like Flux and SD35 add more diversity by default but still show most of the tested biases. The limits of a prompt-engineering approach with LLMs become visible for disabilities. Older models like SDXL tend to reverse the relation e.g., for doctors and have the patient sitting in a wheelchair instead of the doctor. This can be interpreted as a bias, or as bad instruction following. Both SD35 and Flux do not show this limitation, although the images are equally biased with the unmodified prompts (e.g., depicting the doctor as a middle-aged white man in four of four cases). Therefore, the author sees this rather as an instruction following problem.

To make the approach more practically feasible, a smaller LLM should be trained with the examples from Claude and ChatGPT (as far as the license permits that) to add less time to the latency of the overall image generation process and keep resource usage within reasonable limits. Country-specific defaults that refer to the statistical distribution in the specific population and user-specific preferences, as suggested by [8], could be implemented as specific system prompts for the LLMs or user's being able to edit (parts) of the task prompt.

For future research, the usage of a VLM as a judge to look for biases still present and hint towards those to send the user a signal like "I recognize that my outputs are not ideal and hope that this is still acceptable for your" should be investigated. This could simulate a kind of self-reflexivity, that might please end-users that are sensitive to biases. Additionally, the LLM could be asked to generate eight diverse prompts instead of four and the four not used for generating images could be suggested for the end user for adding even more diversity.

What most users presumably not want to read is the analysis of the LLM regarding what biases might arise from a the original prompt, in order not to feel like a pupil being reprimanded by the teacher.

## ACKNOWLEDGMENTS

The work was partially supported by the EFRD project: Multi-modal Man-Machine Interface with AI (M4-SKI)